\documentclass[10pt]{article}
\usepackage[letterpaper,margin=.85in]{geometry}
\usepackage{times,natbib}
\setcitestyle{authoryear,round,citesep={;},aysep={,},yysep={;}}
\usepackage{amsmath,amssymb,booktabs,array,graphicx,microtype}
\usepackage{xcolor}
\usepackage{caption,float,placeins}
\usepackage{hyperref,url}
\definecolor{ScopeBlue}{HTML}{5271AE}
\definecolor{ScopeLight}{HTML}{70ACDE}
\definecolor{ScopeGold}{HTML}{F5CC7D}
\definecolor{ScopeOrange}{HTML}{FFA660}
\definecolor{ScopeRed}{HTML}{D85B59}
\hypersetup{pdftitle={ControlScope: Workflow Revision and Reliability in LLM Agents},pdfauthor={Jingjie Ning, Xueqi Li, Yibo Kong, Dongting Li},pdfkeywords={LLM agents, workflow revision, agent control, tool use, reliability},colorlinks=true,citecolor=ScopeBlue,linkcolor=ScopeBlue,urlcolor=ScopeBlue}
\newcommand{\keep}{\textsc{Keep}}
\newcommand{\argu}{\textsc{Arg}}
\newcommand{\full}{\textsc{Full}}
\newcommand{\scope}{\mbox{\textsc{ControlScope}}}
\title{\textbf{ControlScope: Workflow Revision and Reliability in LLM Agents}}
\author{%
\makebox[\dimexpr\textwidth-2\tabcolsep\relax][c]{%
\begin{tabular}{c}
\textbf{Jingjie Ning}\textsuperscript{1}\thanks{Corresponding author}\quad
\textbf{Xueqi Li}\textsuperscript{1}\quad
\textbf{Yibo Kong}\textsuperscript{1}\quad
\textbf{Dongting Li}\textsuperscript{2}\\[4pt]
{\normalfont\small \textsuperscript{1}Carnegie Mellon University\quad
\textsuperscript{2}Tsinghua University}\\[2pt]
{\normalfont\footnotesize
\{\href{mailto:jening@cs.cmu.edu}{jening},\,
\href{mailto:xueqil@cs.cmu.edu}{xueqil},\,
\href{mailto:yibok@cs.cmu.edu}{yibok}\}@cs.cmu.edu\quad
\href{mailto:ldt25@mails.tsinghua.edu.cn}{ldt25@mails.tsinghua.edu.cn}}
\end{tabular}}%
}
\date{}
\begin{document}
\maketitle

\begin{abstract}
How much of a running workflow should a language model agent revise? \scope\ compares continuing generated code, editing the next tool call's data arguments, and replacing the unfinished workflow from the same public execution state. The nested permissions separate available repairs from the actions an agent selects. We evaluate one-time and repeated reviews across filesystem tasks, ALFWorld, and AppWorld. Across two source programs per task and three reasoning-reviewer draws on 20 filesystem tasks, \full\ completes 15--16 tasks versus 13 for \keep; across four fast draws it completes 10--13 versus 13. Fresh student-record confirmation reproduces a batch-read repair. ALFWorld fast panels yield \keep/\argu/\full\ scores of 85/86/87 on 87 tasks across 52 scenes and 134/134/127 on 134 tasks across four scenes; reasoning on the 87-task cohort also yields 85/86/87 with substantial review cost. An AppWorld V1 official-test panel of 585 task instances from 195 scenario templates shows small net differences. Frozen replays expose viable agent-written replacements interrupted by later revision in two failed file-organization runs. An offline source-trajectory midpoint comparison shows later reviews completing an insufficient repair. Five-call protection saves 19.4\% of logged model output and loses one success across 20 fresh source runs. An argument-only shortcut shows that the broader sampled policy can overlook a cheaper successful edit available in both operation sets. These outcomes tie repair access to actual choices and subsequent execution.
\end{abstract}

\begin{figure}[H]
\centering
\includegraphics[width=\linewidth]{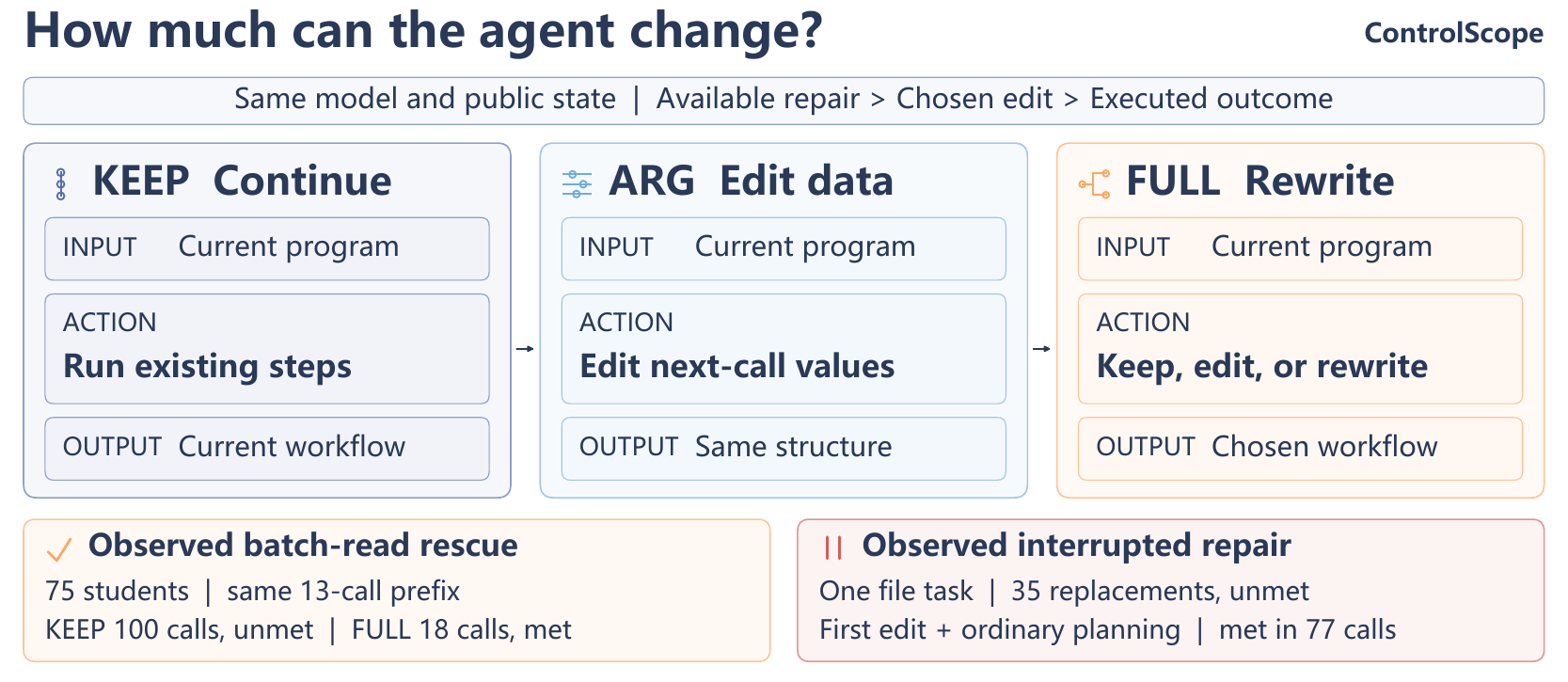}
\caption{\scope\ varies the part of an existing program an agent may revise from the same public state. \full\ includes the smaller actions. The lower cards show a matched batch-read repair and a file-organization replay that retains its first replacement with ordinary planning.}
\label{fig:teaser}
\end{figure}

\section{Introduction}
Language model agents increasingly act through executable workflows. A model may generate a loop over files, a sequence of application calls, or a plan for interacting with an environment. Program execution then carries out decisions that the model has already made. This division of labor supports efficient tool use, while making the authority to revise a running program an important architectural choice \citep{codeact,llmcompiler,llmascode}.

A concrete example illustrates the choice. An agent must inspect student records and produce a contact list. Its generated program reads records individually. Under a fixed tool-call budget, a reviewer can correct the next file path or replace the loop with batched reads. Both reviewers receive the same public execution record and tool descriptions. Their permitted edits determine which repair they can make. Further workflow replacements can interrupt a useful continuation before it~\mbox{finishes its work.}

We ask how much adaptation follows from expanding the scope of executable revision. \scope\ compares three nested conditions, shown in Figure~\ref{fig:teaser}. \keep\ continues the current program. \argu\ also allows edits to the next primitive tool call's data arguments. \full\ additionally allows replacement of the remaining workflow. Here a primitive call invokes an original environment API directly, and a workflow is the program that organizes these calls. \full\ can choose either smaller-scope operation~\mbox{at every opportunity.}

Our experiments reveal useful adaptation and execution disruption. Across two source programs and three reasoning-reviewer draws, \full\ solves two or three more tasks than \keep\ out of 20; four fast draws range from three fewer successes to equal success. The AppWorld V1 official-test panel shows small aggregate differences across 585 task instances from 195 scenario templates. In file organization, frozen replays show that viable continuations can be interrupted by later revisions. At an early single revision boundary, 216 paired draws yield equal \argu\ and \full\ terminal outcomes. Fixed review schedules expose a tradeoff between subsequent repair opportunities and model work. Fresh-subset confirmation and an argument-only shortcut connect available control to actual choices~\mbox{and workload.}

We use three complementary diagnostic dimensions. \emph{Repair availability} has a positive witness when a recorded permissible edit reaches the goal under an explicit continuation protocol. Candidate-only replay tests the edit through its block or call budget; retained-edit replay also allows ordinary planning. \emph{Repair selection} records the chosen edit. \emph{Execution stability} records later reviews and execution. Matched prefixes connect these observations to outcomes.

\section{Related Work}
\label{sec:related}
\textbf{Agents that reason, plan, and execute.}
ReAct interleaves reasoning with environment actions \citep{react}, while Toolformer learns API use through language modeling \citep{toolformer}. Plan-and-Solve structures reasoning into planning and execution stages \citep{plansolve}. Code as Policies generates executable programs for embodied control \citep{codepolicies}, while ReWOO separates reasoning from external observations \citep{rewoo}. Plan-and-Act separates a learned planner from an executor \citep{planact}. CodeAct represents actions as executable Python \citep{codeact}, and LLMCompiler schedules function calls through a compiler-inspired design \citep{llmcompiler}. ReCode recursively decomposes code into actions at different granularities \citep{recode}. LLM-as-Code places control flow in a program and uses the model within that structure \citep{llmascode}. These approaches establish the architectural choices that motivate our comparison. We hold an already-generated program fixed and vary the changes a reviewer can make \mbox{during its execution.}

\textbf{Feedback, repair, and revision.}
Reflexion uses verbal feedback across trials \citep{reflexion}, Self-Refine iteratively improves generated outputs \citep{selfrefine}, and self-debugging uses program execution to support code correction \citep{selfdebug}. AdaPlanner refines plans through feedback within and across plans \citep{adaplanner}; Language Agent Tree Search combines reasoning, action, and search \citep{lats}. GraSP explicitly represents skill dependencies and compares typed local repair with global replanning \citep{grasp}. Its local-repair findings provide a direct precedent for scope-sensitive recovery. Anticipatory reflection explicitly addresses the tension between frequent plan changes and consistent execution \citep{devil}. Neuro-symbolic code validation grounds plans through symbolic checks and environment interaction \citep{reliablecode}; plan--memory coupling addresses repeated failures in software repair \citep{planmemory}. Our comparison measures nested action sets and the agent's actual choices from each set. Work on second-pass revision further shows that improvement depends on the information and structure supplied by the initial draft \citep{revision}. Recent itinerary-revision experiments compare full replanning, hierarchical repair, and local editing while measuring feasibility and plan stability \citep{travelrepair}. We replay selected operations from the same state under a shared model and backend, then observe how the revised continuation~\mbox{executes.}

\textbf{Intervention value and execution interfaces.}
\citet{calibration} evaluate interventions by branching from the same trajectory prefix and distinguish intervention value from continuation risk. DIAL learns when extra computation is beneficial from counterfactual exploration \citep{dial}. We use same-state branching to study the executable scope of a revision and the outcome of the operation an agent selects. Planning-horizon comparisons analyze how often agents return to the model \citep{horizon}. Model Context Protocol (MCP) design studies and enterprise tool-interface comparisons show that execution interfaces affect agent behavior \citep{mcpdesign,bash}. Accordingly, our main comparisons share a tool backend, and our configuration analysis includes a matched-interface fast control. CaMeL separates control flow from untrusted data to enforce security properties \citep{camel}. We study task completion with system permissions held fixed, providing a complementary view~\mbox{of workflow control.}

\textbf{Interactive evaluation.}
AgentBench evaluates language models across interactive environments \citep{agentbench}. ToolSandbox provides stateful tool-use evaluation \citep{toolsandbox}. AppWorld supports interactive coding across application APIs \citep{appworld}, MCPMark supplies realistic tool tasks and executable checks \citep{mcpmark}, and ALFWorld connects language interaction with embodied household goals \citep{alfworld}. Our measured endpoints span MCPMark-derived filesystem tasks, ALFWorld, and an AppWorld V1 official-test panel. We state each execution contract and the filesystem task-selection rules so that scope effects can be interpreted alongside the~\mbox{underlying benchmark contracts.}

\section{Measuring Revision Scope}
\label{sec:design}
\subsection{A shared execution state and nested operations}
Let $h$ denote the complete public execution record available at a decision boundary. It includes the user goal, observed tool results, current program, pending call, and exposed program variables. Let $s$ denote the corresponding environment state. A \emph{continuation} is the code and queued tool calls that have already been generated and remain to execute. Branches begin from the same $(h,s)$ and use the same underlying model, primitive \mbox{APIs, and permissions.}

We define the permitted operation sets by
\begin{align}
\mathcal{A}_{\keep}(h) &= \{\operatorname{keep}\}, \\
\mathcal{A}_{\argu}(h) &= \mathcal{A}_{\keep}(h) \cup \mathcal{P}(h), \\
\mathcal{A}_{\full}(h) &= \mathcal{A}_{\argu}(h) \cup \mathcal{R}(h),
\end{align}
where $\mathcal{P}(h)$ contains valid edits to the next primitive call's data arguments, and $\mathcal{R}(h)$ contains valid replacements of the remaining continuation. Argument revision preserves the pending API and subsequent program structure. Changing a path, identifier, or data value can still change later behavior through the program's existing branches. Workflow revision can change call selection, sequencing,~\mbox{loops, and dependencies.}

The executor exposes explicit operations for keeping, patching arguments, and replacing the continuation. The same patch implementation serves \argu\ and \full. Parameters that directly encode executable programs fall outside the data-argument comparison. Whole-reply \full\ panels cover the current block and later tool calls queued in the same assistant reply. The two P1 fast panels use current-block replacement (Table~\ref{tab:main}).

\textbf{Granted and realized scope.}
Granted scope is the set of operations available to the model. Realized scope is the operation it actually chooses. Inclusion of the smaller sets gives \full\ access to every restricted choice. Consequently, any loss from the larger set reflects the deployed selection and execution policy, including its interaction with the interface and budget. We log both the granted condition and~\mbox{every realized operation.}

We measure \emph{reliability} as native terminal task success under a stated policy and budget. \emph{Execution stability} records whether a selected revision persists through later review and planning.

\subsection{Single revisions and sustained policies}
A \emph{local intervention} grants one revision opportunity at a fixed boundary. Subsequent scope decisions are \keep. A \emph{sustained revision policy} offers the assigned scope repeatedly during execution. These are separate experimental treatments because later revisions can change whether an earlier \mbox{repair reaches completion.}

Every condition retains ordinary planning when a code block or assistant reply finishes. In ALFWorld, every condition also retains the same ordinary failure-recovery procedure. This common planner can produce a new program after observing execution results. The scope comparison therefore measures control over an existing unfinished program within an otherwise~\mbox{functioning agent loop.}

For binary terminal task success $Y$, we report paired differences
\begin{align}
\Delta_{A,K}&=\mathbb{E}[Y(\argu)-Y(\keep)], &
\Delta_{F,A}&=\mathbb{E}[Y(\full)-Y(\argu)], \\
\Delta_{F,K}&=\mathbb{E}[Y(\full)-Y(\keep)].
\end{align}
The paired execution unit is a task--source-program instance. Source programs and reviewer draws from the same task share its structure and remain grouped in interpretation. Filesystem uncertainty resamples the seven task categories.

\subsection{Replay and candidate execution}
A \emph{prefix replay} reconstructs all environment interactions preceding a selected boundary in an isolated workspace. We compare API names, arguments, observations, and terminal files with the recorded source. File-clock fields are handled through an explicit normalization that preserves task-relevant modification times. Hidden evaluation code stays outside the agent workspace. Each selected branch receives the same public record, while later observations evolve with \mbox{its own actions.}

We use two complementary replay diagnostics. First, an actual \full\ argument patch is executed through both scope labels from the same state. This checks that the candidate also belongs to the lower-scope execution path. Second, an actual workflow replacement is retained while subsequent scope interventions are disabled. One endpoint evaluates goal completion by the replacement block or original budget. A separate endpoint evaluates complete agent recovery after the retained edit and common ordinary planning. We identify the endpoint for each~\mbox{reported witness.}

\section{Experimental Setup}
\label{sec:setup}
\textbf{Tasks and source programs.}
An evaluation panel fixes a task set, one source program per task, and a reviewer configuration. A source program is sampled before the scope comparison. The main filesystem comparison contains 20 tasks across seven categories, selected from a 24-task MCPMark adaptation after inspecting instruction constraints. Four tasks restrict Python use and are excluded from this comparison; their scored outcomes remain in the full 24-task supplementary records. The adapter gives generated Python code access to the original filesystem APIs through a proxy. Each task has two independently generated source programs, reflecting prior evidence that implementation choice can change measured outcomes in automated research \citep{onerun}. Native file-target verifiers determine task success under the stated~\mbox{execution contract.}

ALFWorld tests household plans in two path-defined cohorts fixed before policy outcomes. All 87 valid\_seen tasks outside development scenes cover 52 scenes; all 134 valid\_unseen tasks cover four. Both cohorts use fast reviewers with required tool calls and an 8,192-token cap; malformed reviews become plan errors handled by ordinary recovery. The 87-task reasoning-reviewer panel uses automatic tool choice, a 16,384-token cap, and \keep\ continuation after invalid reviews. Scores include~\mbox{common recovery.}

An AppWorld V1 panel adds 585 official-test tasks across 195 scenario templates, paired by initial policy draw. It offers one review per natural block, with a 40-block task cap and 250 API requests per block. Its current-block replacement renews that block's allowance; the whole-reply filesystem panels share a fixed 100-call task cap. We report official-test outcomes as~\mbox{split-level aggregates.}

\textbf{Model and budget.}
Source generation and ordinary planning use served \texttt{deepseek-flash} in fast mode; reviewers use fast or high-effort reasoning mode. Matched configurations share native tool schemas, an explicit tool-return instruction, automatic tool choice, and a 16,384-token per-request cap. Fast sampling uses temperature 0.7; reasoning uses the provider's reasoning-mode settings.

MCP conditions receive 100 primitive calls, 100 ordinary planning turns, and 1,800 execution seconds; review latency is recorded separately. ALFWorld uses 100 actions and up to five ordinary repairs. MCP and ALFWorld reasoning-reviewer errors fall back to \keep; ALFWorld fast review errors enter ordinary recovery. An MCP ordinary-planner error ends the trajectory, with native grading of the existing files. Paired analyses use validated prefixes and~\mbox{worker execution.}

\textbf{Analysis panels.}
We distinguish the completed sustained-policy panels, the repeated local experiment, and task-family follow-ups. The local experiment freezes 36 eligible states from 19 tasks across two source programs. Each state receives three \argu\ and three \full\ draws in each reviewer mode, yielding 432 decisions and 216 paired draws. Four task--program combinations lack a supported boundary. Each task--source pair reuses one cached \keep\ baseline; sampled reviewed continuations share ordinary-planner responses only when their exact requests coincide.

The primary tables report counts so that small denominators remain visible. We retain positive, negative, and equal outcomes. A seven-category bootstrap supplies exploratory uncertainty for the MCP task set; the number of independent categories governs its precision. All task-family follow-ups are identified as generated variants or subsets, and repeated source programs retain their~\mbox{shared task identity.}

\section{Results}
\label{sec:results}
\subsection{Workflow revision provides configuration-dependent gains}
\begin{table}[htbp]
\centering
\caption{Sustained-policy panels on the same 20 tasks. P1/P2 are independent source programs; draws from one source reuse \keep. Replace marks current Block or whole Reply. F W/L counts paired \full\ wins/losses against~\keep.}
\label{tab:main}
\includegraphics[width=\linewidth]{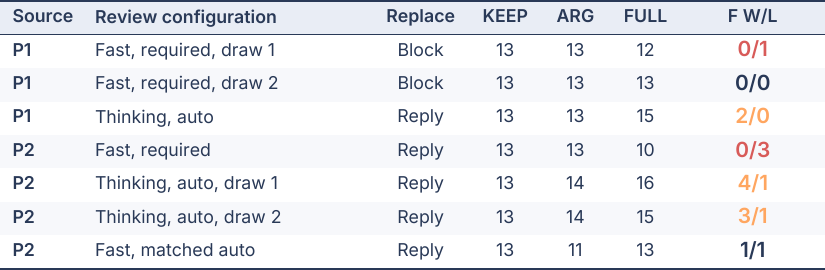}
\end{table}

Table~\ref{tab:main} reports two source programs and three reasoning-reviewer draws. \full\ solves 15--16 tasks versus 13 for \keep. P1 gains contact inference and English-student selection. Both P2 draws gain contact inference, file splitting, and grade scoring; draw 1 also gains English-student selection and loses file arrangement, while draw 2 loses legal solution tracing. Thus, the source program and reviewer draw shape which tasks benefit~\mbox{from revision.}

The matched fast panel completes 13, 11, and 13 tasks under \keep, \argu, and \full. Its equal \keep/\full\ totals contain one gain in grade scoring and one loss in legal solution tracing. Relative to \argu, \full\ gains grade scoring and a music-report task. The \argu/\full\ reviewers record 29/16 rejected proposals, which remain in the scored policy outcomes.

The fast P2 required-tool configuration scores 13/13/10 for \keep/\argu/\full, while automatic-tool fast scores 13/11/13. P2 provides a whole-reply, automatic-tool comparison across fast and reasoning reviewers; P1 fast also changes replacement boundary. Forty P2 first requests match in messages, schemas, tool choice, token cap, and model identifier. Reasoning settings and stochastic draws differ. The seven-category bootstrap 95\% intervals for P2 \full-\keep\ are $[-30.0,-4.2]$ and $[-14.3,16.7]$ percentage points for required-tool and automatic-tool fast, and $[0,38.9]$ and $[-10.0,27.8]$ for the two reasoning draws. Full per-row intervals are in the supplement.

\begin{figure}[htbp]
\centering
\includegraphics[width=\linewidth]{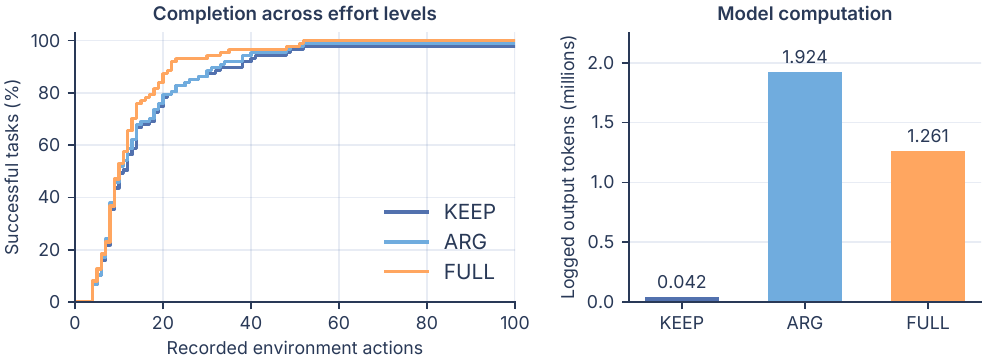}
\caption{Recorded ALFWorld effort and model computation. The left panel gives the percentage of tasks completed within each environment-action threshold under the shared 100-action budget. The right panel reports logged output tokens, including reused requests. Both include~\mbox{common recovery.}}
\label{fig:scope}
\end{figure}

\subsection{Common recovery changes the visible benefit}
On the 87-task reasoning-reviewer panel, initial-plan successes are 79, 81, and 85 for \keep, \argu, and \full. After ordinary recovery, these become 85, 86, and 87. Figure~\ref{fig:scope} shows execution effort. Within 20 actions, completion percentages correspond to 68, 69, and 76 tasks. \full\ uses 1,124 actions versus 1,336 for \keep\ and invokes ordinary repair twice versus 17 times. Logged input tokens total 0.221, 15.926, and 11.881 million; exact output totals are 42,481, 1,924,261, and 1,261,096 tokens. \argu/\full\ record 51/16 invalid review replies, retained in task scores. \full\ gains two terminal successes over \keep\ with 29.7 times its output tokens. Its 34.5\% lower total output than \argu\ concentrates 98.3\% in two heating tasks with 28 invalid \argu\ reviews and none under \full; \full\ logs more output on 60 of 87 tasks. These configuration totals reflect reviewer validity and subsequent workload (Appendix~\ref{app:alfresources}).

On the 87-task cohort, fast and reasoning reviewers both end 85/86/87. Under the same fast protocol, the 134-task valid\_unseen cohort ends 134/134/127 across four scenes. The observed \full\ direction changes across these fixed cohorts (Appendix~\ref{app:alfresources}).

Within the 87-task reasoning-reviewer panel, two heating tasks account for all terminal differences. In scene 28, \full\ replaces an unsuccessful stove-burner procedure with a microwave sequence after the same four initial actions and first recovery input. In scene 1, both \argu\ and \full\ recover; \argu\ combines argument edits with the common planner. \argu\ records 15 and 13 invalid reviews in these two tasks, while \full\ records~\mbox{none.}

\subsection{AppWorld official tests show small aggregate differences}
\begin{table}[htbp]
\centering
\caption{AppWorld V1 official-test aggregates. Task goal completion (TGC) counts successful tasks; scenario goal completion (SGC) is the percentage of templates with all three instances successful. K/A/F denote \keep/\argu/\full.}
\label{tab:appworld}
\includegraphics[width=\linewidth]{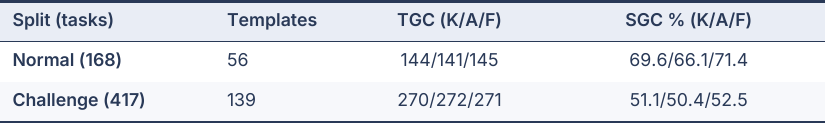}
\end{table}

Across 1,755 completed AppWorld V1 runs, \full\ exceeds \keep\ by one task on each split (Table~\ref{tab:appworld}). Paired TGC gains are $+0.60$ and $+0.24$ percentage points; scenario-cluster bootstrap 95\% intervals are $[-2.38,3.57]$ and $[-2.40,2.88]$. Paired \full\ win/loss counts of 4/3 and 18/17 reveal changes in both directions. Each V1 \full\ replacement also renews a 250-request block allowance. Its observed difference combines revision scope with this extra capacity, which favors~\full.

A separate V2 local diagnostic covers 86 development boundaries from 28 templates, with two draws and four permissions per boundary. \keep, \argu, coordinated data editing (\textsc{Params}), and \full\ each succeed on 154/172 continuations; \full\ selects 10 workflow replacements. V2 uses shared remaining-block quota and a private-runtime guard; \full\ includes every \textsc{Params} edit. In one pagination scenario, fixed \keep, next-call edit, \textsc{Params}, and workflow candidates succeed 0/3, 2/3, 3/3, and 3/3; online \textsc{Params} and \full\ selection succeed 0/3 and 1/3 (Appendix~\ref{app:appworld}).

\subsection{A single early intervention often leaves outcomes unchanged}
All 216 paired local draws have identical \argu\ and \full\ terminal outcomes. Two reasoning-enabled draws on the second file-splitting source improve over \keep, and both scopes succeed in those draws. The other paired draws match \mbox{the baseline outcome.}

Realized choices help explain this pattern. Of 432 decisions, 391 ultimately continue unchanged, including 13 malformed responses and 12 service-error fallbacks. The remaining decisions contain 18 argument patches and 23 workflow replacements. The 12 service errors all arise from a single oversized context; excluding that state preserves the zero observed \argu--\full\ difference. All 41 actual revision branches pass the saved-context, prefix, operation,~\mbox{and execution-budget checks.}

The paired equality describes a fixed early boundary in 19 distinct tasks. Sustained policies reach additional states and can exercise different~\mbox{revision opportunities.}

\textbf{Visible runtime feedback.}
A paired information diagnostic gives both reviewers \full\ authority once at the same natural execution boundary. A code-and-history view contains the generated program and prior public history; a feedback view adds current-block receipts, pending arguments, and runtime values. Across 17 eligible pairs covering 11 tasks, each view succeeds on 12/17, with 11 joint successes. Feedback selects nine replacements versus three, with one paired gain and one loss. This measures incremental visibility at the paused review (Appendix~\ref{app:feedback}).

\section{How Revision Helps and Hurts}
\label{sec:mechanisms}
\subsection{Available repairs and selected edits}
\textbf{Batching reads.}
The English-student selection task requires combining basic student information with teacher recommendations. In one original source, a retained batch-read replacement reaches the goal after 94 total calls, whereas continuing the original loop reaches the 100-call budget without completing the task. Both branches share the first 88 calls. A second source yields the same mechanism at an earlier boundary, with the replacement completing the task at 28 total calls from an identical 26-call prefix. The resulting 19 selected students are independently checked against all original records. These frozen comparisons use recorded model choices and introduce no \mbox{new review sampling.}

\textbf{Repairing a parser.}
Some student CSV records contain unquoted commas inside addresses. A workflow replacement can revise the parsing logic around reliable fields and produce correct results for all 150 students. The reasoning configuration succeeds repeatedly on the affected source, and the matched fast configuration also succeeds. The other natural source already parses the records correctly. This contrast ties the gain to a repairable property of the initial program. Some unsuccessful argument-review draws exhaust their output budget before returning a valid operation, and the recorded outcomes~\mbox{include those fallbacks.}

\textbf{Workload follow-up.}
Across 25-, 75-, and 150-student subsets with two seeds and natural source programs, both reviewer modes score 5/6, 5/6, and 6/6 for \keep/\argu/\full. The gain occurs on one 75-student seed-1 source. There, \keep\ and \argu\ exhaust 100 calls; fast \full\ switches to batch reads after an identical 32-call prefix and succeeds in 42 calls, while reasoning \full\ succeeds in 10. Fast \argu\ selects three valid patches. Both 150-student sources already succeed under \keep. The generated program's reading strategy shapes the observed revision opportunity.

\textbf{Confirmation on new subsets.}
We select the discriminating 75-student setting after inspecting the workload exploration, then evaluate three new subsets with two independently generated source programs each. Both reviewer modes again yield 5/6, 5/6, and 6/6 successes (Table~\ref{tab:confirmation}). On seed 101's first source, \keep\ and \argu\ exhaust 100 calls; \full\ succeeds in 18 fast-mode calls and 15 reasoning-mode calls. The fast repair follows 13 identical API calls and receipts, with zero rejected proposals or service errors. The second source already succeeds under \keep\ on all three subsets. These comparisons preserve source-program dependence within the~\mbox{same task family.}

\begin{table}[htbp]
\centering
\caption{Fresh-subset confirmation on six task--program pairs per mode. Triplets follow \keep/\argu/\full. Calls and logged output include reused requests; output is shown in thousands to two decimals. KEEP is~\mbox{shared between modes.}}
\label{tab:confirmation}
\includegraphics[width=\linewidth]{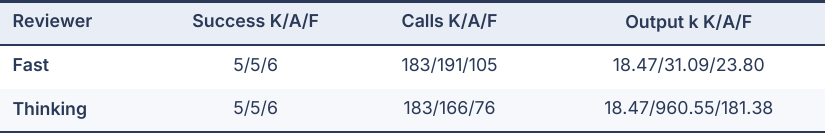}
\end{table}

The confirmation also connects revision scope to computation. Relative to \argu, \full\ uses 23.4\% fewer logged output tokens in fast mode and 81.1\% fewer in reasoning mode, computed from unrounded totals. A selected rewrite changes later tool workload, while reasoning \argu/\full\ record 24/1 invalid-review fallbacks. Both contribute to the logged token contrast.

\textbf{An argument-only efficiency shortcut.}
On seed 102's first source, the pending API supports batched input. The reasoning \argu\ policy expands its path list from 2 to 75 and succeeds in 11 calls. \keep\ and the sampled \full\ policy also succeed, each using 25 calls. The public state at the sixth-call boundary matches between \argu\ and \full, which chooses to continue unchanged. Shared-patch replay succeeds under both labels in 11 calls with matching traces and files. The sampled \full\ reviewer leaves this shortcut unused, adding 14 primitive calls while~\mbox{preserving success.}

\subsection{Later reviews shape repair completion}
A successful candidate can lose its opportunity to finish when later reviews keep replacing it. We examine this behavior in the file-time classification task, where files must be moved into date-based directories and accompanied by metadata. The original execution encounters a genuine missing-parent-directory error. The agent's ensuing rewrite attempts address a \mbox{real recovery need.}

\begin{figure}[htbp]
\centering
\includegraphics[width=\linewidth]{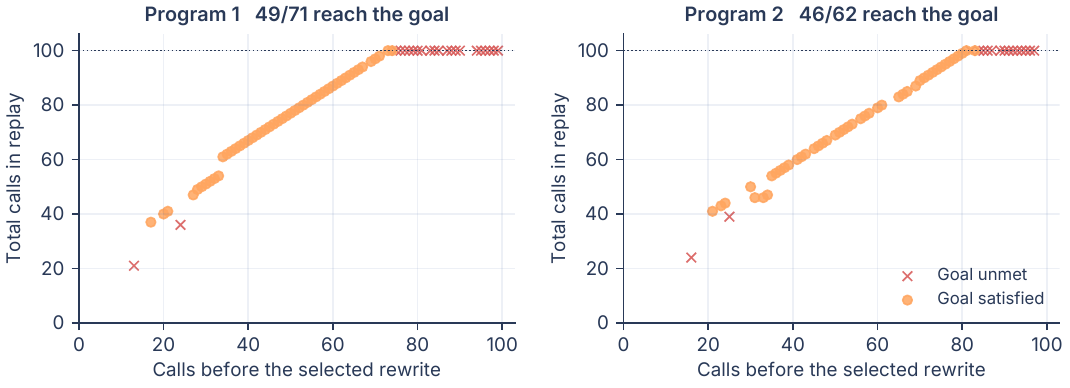}
\caption{Frozen execution of every actual replacement in two failing runs of file-time classification. Program 1 uses P1 fast required draw 1; Program 2 uses P2 fast required. Each point retains the recorded prefix and executes the selected block to completion or the original budget. Goal satisfaction occurs for 49/71 and 46/62 candidates from one task with correlated prefixes.}
\label{fig:rewrite}
\end{figure}

For the first source, 49 of 71 actual replacements satisfy the task goal when retained through the block endpoint or the original call budget. For the second source, 46 of 62 do so (Figure~\ref{fig:rewrite}). Forty-eight of the first source's successful candidates also finish their replacement block. These are correlated candidate-level replays within two runs of one task. They establish that useful continuations already existed within the failing~\mbox{policies' own outputs.}

We also generate a small file-organization workload grid and select its discriminating setting for fresh data seeds. In three confirmation instances, \keep\ and \argu\ succeed on all three, while \full\ succeeds on two. The failed run makes 35 replacements and reaches the 100-call budget. Retaining its first replacement and then allowing common ordinary planning succeeds in 77 calls. The first replacement alone ends after seven total calls with the goal still incomplete. This separates the value of preserving subsequent execution and planning from the ability of one code block to solve~\mbox{the entire task.}

A broader retained-first-revision comparison covers every replacement-producing run in the P2 automatic-tool fast panel and first P2 reasoning draw. Sustained revision succeeds on 5/10 fast and 9/13 reasoning runs; retaining only the first revision yields 4/10 and 7/13, with zero paired gains and three losses. Later reviews complete repairs in those three changed outcomes. The interruption evidence remains case-level; this tested retention rule yields no panel-level gain.

\textbf{Source-trajectory-defined midpoint schedule.}
The offline comparison uses P2 source programs and fixes each first-review boundary from the completed source trajectory length. This rule requires the baseline trajectory and serves as a paired diagnostic. One-time and continuous reviews share the first decision within each scope, and ordinary planning remains common.

Table~\ref{tab:schedule} shows 12/20 successes under either \argu\ schedule, while \full\ succeeds on 13/20 with one review and 14/20 with continuous review. The sole within-scope terminal change is English-student selection. After the same first \full\ replacement, continuous review makes three later replacements and succeeds in 58 calls; one-time review exhausts 100 calls. The continuous \full\ policy uses 294 reviews against 20 and logs 64.0k against 45.8k output tokens.

\begin{table}[htbp]
\centering
\caption{Source-trajectory-defined midpoint schedule on 20 matched fast filesystem tasks. The shared \keep\ baseline succeeds on 13/20. Edits count accepted patches and replacements; output includes reused requests. One-time legal \argu\ is scored after an ordinary-planner HTTP~400.}
\label{tab:schedule}
\includegraphics[width=\linewidth]{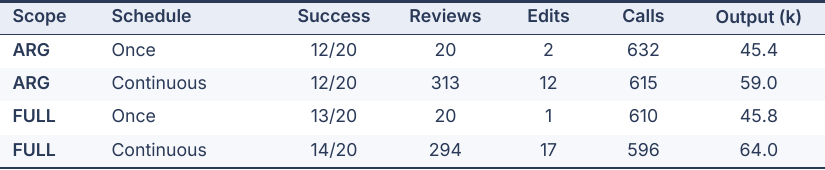}
\end{table}

Across 20 fresh source programs, \keep/\argu/continuous-\full/protected-\full\ succeed on 13/12/14/13 tasks. Five-call protection activates on eight tasks, with zero paired wins, one loss, and 19 equal outcomes against continuous \full. Reviews fall from 517 to 423 and logged output falls 19.4\%. In the sole flip, continuous \full\ selects a 300-path batch read and succeeds; protected \full\ omits it and exhausts 100 calls (Appendix~\ref{app:protect5}).

\section{Implications for Agent Design}
\label{sec:implications}
Agent systems should expose both data edits and workflow replacement, since batch-read and parser repairs require wider control while a shared argument edit can save 14 calls. Review timing and valid-operation return rates belong in the same evaluation. Five-call protection saves 19.4\% of output while losing one success.

\FloatBarrier
\section{Conclusion}
\scope\ compares \keep, \argu, and \full\ revisions from matched public states. Across two source programs per task and three reasoning-reviewer draws, \full\ completes 15--16/20 tasks versus 13/20 for shared \keep; four fast draws give 10--13/20 versus 13/20. Rewrites repair batching and parsing, with fresh student-record confirmation. A shared edit saves 14 calls, while frozen replays expose interruption in two runs of one file task. Both tested retention rules have zero paired gains. Retaining the first edit loses 3/23 replacement-producing runs, and five-call protection loses 1/20 tasks while saving 19.4\% of output. \argu\ and \full\ match in all 216 early local pairs; AppWorld V1 has small net differences with a quota favorable to \full. Under a common fast ALFWorld protocol, \full\ gains two tasks across 52 scenes and loses seven across four scenes. Edit scope, valid selection, and review timing jointly shape task success and model work.

\section*{AI Use Statement}
Generative AI assistants contributed to the research question, experimental design, code implementation, result analysis, literature review, manuscript writing, and figure production. Their outputs were checked against saved execution records, native task scores, and bibliographic sources. The authors take responsibility for the resulting claims and artifacts.

\clearpage
\bibliography{references}
\bibliographystyle{iclr2027_conference}
\clearpage
\appendix
\section{Execution Contract and Reproducibility}
\label{app:contract}
\subsection{Scope boundaries and common planning}
The filesystem agent emits Python blocks whose external effects occur exclusively through a proxy to the original Model Context Protocol (MCP) filesystem server. MCP is the protocol that exposes the tools and their input schemas. Python executes in an isolated worker with no direct task-filesystem mount or network access. The environment server receives a separate task workspace for each branch. Standard-library computation can \mbox{transform returned data.}

At each supported boundary, the worker exposes the pending primitive call, its arguments, JSON-serializable local and module variables, and printed output. The reviewer also sees the task history, current block, public tool schemas, earlier review decisions, and queued calls. A replacement cancels the pending call and substitutes new code for the unfinished assistant reply. Cancellation receipts preserve the tool-call protocol, and the ordinary model~\mbox{planner resumes afterward.}

The two P1 fast runs permit current-block replacement. The P1 reasoning, P2, and local panels permit whole-reply replacement, including queued tool calls. A project-management source has a queued block after the reviewed block; the compared policies choose \keep\ at that boundary and fail the task. Each panel is interpreted under its recorded~\mbox{replacement boundary.}

\subsection{AppWorld V1 protocol}
\label{app:appworld}
The AppWorld V1 official-test comparison uses the same non-reasoning served model, public API descriptions, 16,384-token generation cap, and 40-code-block task cap across its three scopes. Its review occurs once per natural block. A \full\ replacement edits the unfinished current block and starts a new 250-request block allowance; this extra capacity favors \full\ in V1. The V2 local protocol shares the current block's remaining allowance, guards private-runtime access, and permits coordinated data edits. Its \full\ action set includes \keep, next-call editing, coordinated block editing, and workflow replacement. The filesystem protocol cancels queued calls in the same assistant reply and enforces a fixed 100-call task budget. Each panel follows its stated~\mbox{execution contract.}

The official-test report contains 168 normal and 417 challenge tasks, with three task instances for each of 195 scenario templates. All 1,755 runs have final native scores and no outstanding infrastructure failure. SGC counts templates with three successful instances. Paired confidence intervals resample scenario templates. The frozen aggregate records task outcomes, API attempts, protected-criterion failures, review fallbacks, and quota rejections. Mechanism analysis uses development data and the official-test presentation uses the aggregate split-level results~\mbox{in Table~\ref{tab:appworld}.}

A separate V2 local-intervention diagnostic covers 86 eligible natural development boundaries from 28 scenario templates, with two reviewer draws and four conditions for 688 completed continuations. \keep, \argu, \textsc{Params}, and \full\ each succeed on 154/172 local runs. \full\ selects 10 replacements that change API use while terminal success remains equal across conditions. V2 uses shared remaining-block allowance, a private-runtime guard, and coordinated data editing. These outcomes describe local continuations; V1 measures sustained full-task policies.

A fixed-candidate pagination diagnostic uses one AppWorld development scenario and its native evaluator. Across three repeated continuations, \keep\ succeeds 0/3, a next-call page-limit edit 2/3, \textsc{Params} edits to three data fields 3/3, and a fixed paging workflow 3/3. The candidates are frozen before evaluation. Separate online selection runs succeed 0/3 with \textsc{Params} and 1/3 with \full. This case witnesses available edits and incomplete selection in one app-API scenario.

\subsection{Feedback visibility at a fixed boundary}
\label{app:feedback}
For each of the 20 main filesystem tasks and two natural source programs, we freeze the first same-block boundary following an information-returning API call. The pending call must be a unique direct API site outside control-flow constructs. This source-based rule yields 17 eligible task--program pairs from 40 candidates, covering 11 distinct tasks. Only one candidate would qualify if the current block also had to be the task's first block. The selected pairs share environment state, execution prefix, remaining call budget, reviewer model, and a single \full\ review opportunity. Later scope decisions are \keep, while ordinary block-end planning stays available.

The code-and-history view (\texttt{CODE\_ONLY}) contains the public history present when the current assistant reply was generated, its code and queued calls, API documentation, and the pending method and source line. The feedback view (\texttt{WITH\_FEEDBACK}) adds current-reply tool messages, current-block receipts and printed output, the pending call's realized arguments, and public runtime variables. Both review requests use the same operation schema. Earlier blocks' observations can influence the common source program, while feedback-view module variables can carry earlier values. This comparison varies current-block runtime visibility while holding revision authority fixed.

All 17 pairs pass view-mask, exact-prefix, model-request, and native-execution checks, with no review or service fallback. Program 1 has eight eligible pairs, with code-and-history versus feedback success of 7/8 versus 6/8, replacement counts of 1 versus 4, and 215 versus 320 tool calls. Program 2 has nine pairs, with success of 5/9 versus 6/9, replacement counts of 2 versus 5, and 251 versus 324 calls. Each view succeeds on 12/17, with 11 joint successes, one feedback win, one loss, and 15 equal outcomes. Feedback selects nine replacements versus three and accumulates 644 versus 466 tool calls; two grade-scoring source units account for 167 of the 178 additional calls.

In program 1's grade-scoring task, code-and-history \keep\ succeeds in 11 calls. Feedback selects a replacement whose CSV parser misreads unquoted address commas, and common planning exhausts the 100-call budget without producing the target files. In program 2's file-splitting task, feedback succeeds in 29 calls against a failed 33-call continuation under the code-and-history view. Its replacement repeats the pending file-information call, while subsequent ordinary planning writes the split files to the required directory. The two terminal flips thus involve the selected operation and the downstream continuation~\mbox{together.}

\subsection{Runtime and information checks}
The MCP source programs replay in independent workspaces. API arguments and non-clock observations must match before an intervention. File-creation and access times receive explicit replay handling; task-relevant original modification times remain preserved. When execution has already written a file, the corresponding output modification-time comparison is tracked separately. Raw clock differences remain available in \mbox{the execution records.}

Across 40 P2 matched-fast and reasoning first requests, messages, schemas, tool choice, token cap, and model identifier coincide. The P1 first-boundary queue is empty on all 20 tasks. The 174 ALFWorld policy branches have identical first public inputs to their~\mbox{corresponding baseline branches.}

A malformed review response or review-stage provider error produces a recorded \keep\ fallback, after which the existing program continues. A provider error in ordinary planning ends the trajectory, and the native verifier scores the files produced so far. Paired branches pass worker-transport, source-observation, and prefix checks. Fixed worker hash seeds reproduce source set displays; trajectory continuation preserves saved requests, responses, contexts, and decisions.

Fresh workload confirmation uses seeds 101--103, new source programs, and separately sampled~\mbox{review decisions.}

\subsection{Budgets and model requests}
\begin{table}[htbp]
\centering
\caption{Shared model settings and execution budgets.}
\includegraphics[width=\linewidth]{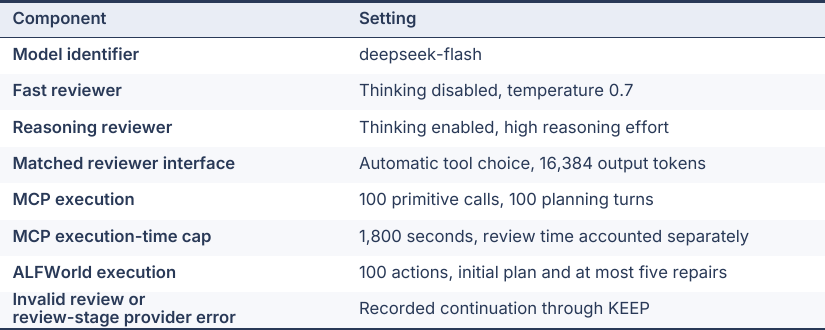}
\end{table}
Exact-request caches couple ordinary planning draws within a paired panel where requests coincide. Review draws remain separately sampled. Reused \keep\ trajectories and cached model responses are identified in resource accounting. Logical token totals describe the model work represented by the trajectories; successful uncached request totals describe a narrower component of actual API use. Monetary cost depends on provider pricing \mbox{and cache treatment.}

\section{Study Coverage and Outcome Accounting}
\label{app:coverage}
The completed main filesystem panels reuse the same 20 tasks. Additional reviewer draws and source programs provide repeated measurements on those tasks. The four instruction-incompatible tasks are pattern matching, structure analysis, structure mirroring, and duplicate student names. Their adapted outcomes remain in the full 24-task record. The seven-category grouping is desktop tasks, desktop templates, file context, folder structure, legal documents, papers, and the student database. Code-repository tasks are kept outside the data-argument comparison because code-valued parameters can expand the \mbox{effective revision set.}

\begin{table}[htbp]
\centering
\caption{Local revision decisions. Each row contains 108 draws across the 36 frozen states. KEEP counts~include~recorded~fallbacks.}
\label{tab:local}
\includegraphics[width=\linewidth]{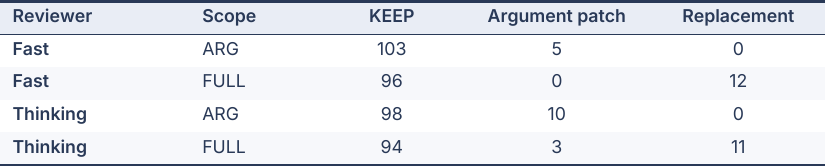}
\end{table}
The local boundary rule chooses the first supported module-level internal boundary of the earliest multi-call writer block, with a read-block fallback. Function- and class-internal boundaries are excluded by this rule. All selected contexts have empty queued-call lists. Four task--program combinations are ineligible, leaving 19 distinct tasks represented by 36 states. Twelve context-limit failures belong to the same second-source paper-organization state. Excluding its draws preserves the observed equality of \argu\ and \full\ outcomes.

\subsection{Source-trajectory-defined midpoint schedule}
\label{app:schedule}
The matched fast schedule experiment uses all 20 tasks and the P2 natural source program for each. The completed source trajectory supplies its total $N$ public API calls. First review then follows $\lfloor N/2\rfloor$ calls; all 20 boundaries fall within the 100-call budget. The boundary is fixed before intervention runs, making this an offline paired diagnostic whose position requires the completed baseline trajectory. Both scopes share the public prefix, and each scope's one-time and continuous conditions replay the exact first review request, response, and decision. Continuous review can recur after a further public call; one-time review makes later scope decisions \keep. Ordinary planning remains available. Five planner-stage HTTP 400 trajectories receive failing native scores; all 80 assigned conditions remain in the analysis.

The \argu\ conditions choose 2 and 12 patches under one-time and continuous review. The \full\ conditions choose one replacement and 16 replacements plus one patch. In English-student selection, both \argu\ schedules and one-time \full\ exhaust 100 calls. Continuous \full\ makes four replacements, performs two batched reads of 150 paths each, and succeeds in 58 calls after ordinary planning writes the 19-student output. Both \full\ schedules share the first replacement. This is the only within-scope terminal change. On legal solution tracing, continuous \argu\ changes agreement-version arguments and produces an incorrect CSV; both \full\ schedules keep the source procedure and succeed. One-time \argu\ completes its scope review and then receives an HTTP 400 from ordinary planning after 36 public calls. The run ends without an output CSV. Its failure includes this planner-stage service event, while continuous \argu\ produces an incorrect CSV without a provider error.

Nine conditions record provider HTTP 400 responses. All four author-folder and four legacy-paper conditions have review-stage errors that fall back to \keep. The legacy-paper runs still succeed; the author-folder runs also terminate after an ordinary-planning HTTP 400 and fail. One-time legal \argu\ has only a planner-stage HTTP 400. Three conditions exhaust the MCP call budget in English-student selection. All outcomes remain in the 20-task denominator for each assigned condition. Exact prefix checks cover the first review and continuation until continuous review gains its next opportunity.

\subsection{Five-call execution protection}
\label{app:protect5}
A second prospective panel generates one new natural source program for each of the same 20 instruction-compatible filesystem tasks before running any policy condition. It compares \keep, \argu, continuous \full, and \full\ with five completed public calls protected after each selected workflow replacement. This rule uses observed call counts during execution. Ordinary block-end planning remains available. All conditions share the 100-call and 100-planning-turn budgets. The first selected replacement and every earlier review decision match exactly between the two \full\ schedules. The scheduler pilot is excluded from the formal panel. All 80 assigned policy runs reach native scores without a technical failure.

The respective \keep/\argu/continuous-\full/protected-\full\ success counts are 13/12/14/13. Protection activates on eight tasks and yields zero paired wins, one loss, and 19 equal outcomes. Continuous and protected \full\ respectively use 537/551 primitive calls, 517/423 scope reviews, 40/30 replacements, 68,025,157/46,245,343 logged input tokens, and 79,075/63,731 logged output tokens. Logical token totals include replayed responses. English-student selection supplies the only terminal change. Both \full\ schedules share a first replacement after seven calls. Continuous \full\ selects a 300-path \texttt{read\_multiple\_files} call at call 88 and writes the target answer at call 89, succeeding in 90 calls. Protected \full\ selects no batch read and reaches the 100-call budget. Provider and budget failures remain in the assigned denominator.

The workload confirmation fixes 75 students across new subset seeds 101--103 and two new natural programs each. These six task--program pairs share the original 150-record pool. All six source scores, 24 scope conditions, and first public inputs pass validation. The confirmation runs use separate source programs and reviewer decisions. Both exploration and confirmation retain every~\mbox{outcome direction.}

The file-organization exploration spans eight generated instances and yields 8/8, 8/8, and 7/8 successes. Three fresh seeds at 16 files and three directory levels yield 3/3, 3/3, and 2/3. Both sets remain grouped within their~\mbox{respective task families.}

\section{Mechanism Diagnostics}
\label{app:mechanism}
\subsection{Shared argument operations}
Every effective \full\ parameter candidate in the audited main panels is replayed under \argu\ and \full\ labels from its recorded prefix. The selected candidate can follow earlier workflow replacements. Its paired execution checks isolate the shared operation implementation. They preserve the distinction between a candidate reachable from the recorded \full\ history and an independently sampled \argu\ policy. The three \full\ parameter candidates in the local-repeat experiment also yield matching API traces, file bytes, and scores \mbox{under both labels.}

The confirmation replays the actual \argu\ patch from seed 102's first source under both labels at the sixth-call boundary. Later scope decisions are \keep\ and ordinary planning remains available. Both replays succeed in 11 calls with matching traces and files. The sampled \keep, \argu, and \full\ policies all succeed; the patch reduces calls from 25 to 11 and witnesses an efficiency~\mbox{selection gap.}

Confirmation invalid-review fallbacks are 2/2 for fast \argu/\full\ and 24/1 for reasoning. All remain in the scored outcomes; service-error fallbacks are zero.

\subsection{Preserving a selected replacement}
The candidate-level file-organization replay in Figure~\ref{fig:rewrite} freezes each actual revision and executes its replacement block within the remaining original budget. Goal satisfaction is checked separately from normal block termination. This distinction accounts for the first source's one successful candidate that reaches the goal at the budget without \mbox{finishing the block.}

A commitment diagnostic retains the first actual replacement and common ordinary planning while disabling later scope changes. All three generated confirmation instances succeed in that condition. The failed sustained-policy instance is rescued in 77 calls; its replacement block alone stops after seven calls with an incomplete goal. Later ordinary planning completes~\mbox{the repair.}

The first-revision comparison includes all 10 fast automatic-tool P2 runs and 13 runs from the first P2 reasoning draw that replace a workflow. All 23 prefixes pass replay checks; later scope decisions are \keep\ while ordinary planning remains available. Fast success changes from 5/10 to 4/10, and reasoning success from 9/13 to 7/13, with zero paired gains and three losses. The losses concern grade scoring in both modes and English-student selection in reasoning mode. Retained author-folder runs end at the model context limit. Exact-request caching couples ordinary planning where requests coincide. Capacity terminations, equal outcomes, and losses remain scored.

\subsection{ALFWorld recovery and computation}
\label{app:alfresources}
Both ALFWorld cohorts were frozen from official paths before policy outcomes. The 87-task valid\_seen cohort excludes 24 development scenes and covers 52 scenes; all 134 valid\_unseen tasks cover four. Under the same fast protocol, \keep/\argu/\full\ score 85/86/87 and 134/134/127 respectively. The 87-task reasoning-reviewer configuration also scores 85/86/87 on the same source plans and budgets. It uses automatic tool choice, a 16,384-token review cap, and \keep\ continuation after invalid reviews. We detail this panel for its 52-scene coverage and complete review-outcome accounting. The fast 134-task cohort supplies the opposite direction.

The 87-task reasoning-reviewer panel has 26 effective \argu\ patches, 31 \full\ replacements, and one \full\ argument patch. The shared patch passes action-trace and native-score replay. Invalid reviews total 51 for \argu\ and 16 for \full; \full\ also has one service-error fallback. All 261 scored trajectories meet the action and repair budgets. Scene 28 has 15 \argu\ fallbacks and zero \full\ fallbacks, retained in the~\mbox{observed configuration effect.}

\begin{table}[H]
\centering
\caption{Recorded outcomes and resources on the 87-task ALFWorld reasoning-reviewer panel. Token totals include reused model responses and describe logged work. The same environment-action and recovery budgets apply across all three scopes.}
\label{tab:alf_resources}
\includegraphics[width=\linewidth]{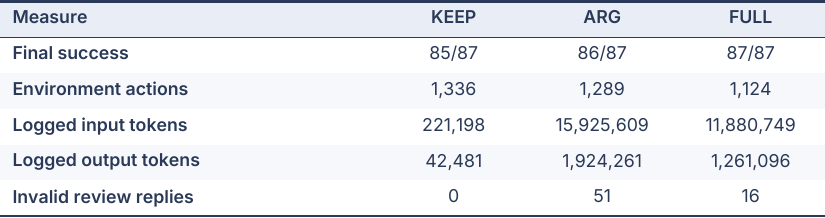}
\end{table}

\end{document}